# Weakly-supervised causal discovery based on fuzzy knowledge and complex data complementarity

Wenrui Li, Wei Zhang, Qinghao Zhang, Xuegong Zhang and Xiaowo Wang*

*Abstract*—Causal discovery based on observational data is important for deciphering the causal mechanism behind complex systems. However, the effectiveness of existing causal discovery methods is limited due to inferior prior knowledge, domain inconsistencies, and the challenges of high-dimensional datasets with small sample sizes. To address this gap, we propose a novel weakly-supervised fuzzy knowledge and data co-driven causal discovery method named KEEL. KEEL adopts a fuzzy causal knowledge schema to encapsulate diverse types of fuzzy knowledge, and forms corresponding weakened constraints. This schema not only lessens the dependency on expertise but also allows various types of limited and error-prone fuzzy knowledge to guide causal discovery. It can enhance the generalization and robustness of causal discovery, especially in high-dimensional and small-sample scenarios. In addition, we integrate the extended linear causal model (ELCM) into KEEL for dealing with the multi-distribution and incomplete data. Extensive experiments with different datasets demonstrate the superiority of KEEL over several state-of-the-art methods in accuracy, robustness and computational efficiency. For causal discovery in real protein signal transduction processes, KEEL outperforms the benchmark method with limited data. In summary, KEEL is effective to tackle the causal discovery tasks with higher accuracy while alleviating the requirement for extensive domain expertise.

## I. INTRODUCTION

CAUSAL relationships are the key factors that determine the behavior and evolution of systems [1], [2]. From the interactions of microscopic particles to the development of human society, causal relationships are interwoven throughout all natural and human activities. Causal discovery refers to the identification of causal relationships between entities within a system using observational data, which is crucial for understanding and intervening in the functioning of systems [1], [3]. In contrast to correlation analyses, causal discovery aims to eliminate confounding effects, elucidating the mechanisms and dependencies behind the observed data [4], [5]. It reveals how changes in one variable can affect others, thereby aiding prediction, intervention, and decision-making across various fields such as biology [6], medicine [7], economics [8], social sciences [9].

### A. Prior Work

Conventional causal discovery methods can be mainly categorized into four groups [1], [10], including constraint-based methods, score-based methods, hybrid methods and function-based methods. All of them start from a foundational assumption that the causal structure corresponds to a directed acyclic graph (DAG), and aim to discover the DAG from the observational data [11], [12].

Constraint-based methods uncover causal DAGs by analyzing conditional independence through statistical tests [1]. For instance, the method CouplaPC is based on Gibbs sampling that integrates the Gaussian Copula model with the PC algorithm [13]. CouplaPC can effectively handle mixed data and increase the reliability of causal discovery. Another method LatentPC combines the mixed latent Gaussian Copula model with the PC algorithm [14], which can effectively identify causal structures in mixed data. However, constraint-based methods require high precision in conditional independence testing which is hard to achieve [15]. As the dimensionality of the data increases, the computational complexity of conditional independence tests also rises [16].

Score-based methods find the optimal graph by maximizing a scoring function that measures how well the graph fits the observed data [1]. These methods involve searching through a space of possible graphs to identify the one with the highest score. For example, CGGES is a causal discovery method based on conditional Gaussian scoring [17], while GGES is a causal discovery method based on generalized score functions [18]. However, these methods have a vast search space, which often requires heuristic search algorithms which are easily trapped into local optimization. Additionally, both the constraint-based and score-based methods fail to distinguish between the Markov equivalence class [19].

Hybrid methods combine the constraint-based and score-based approaches with the aim of improving the accuracy and computational efficiency [1]. For example, HCM employs a score-based method called CVMIC and a constraint-based method named MRCIT, enabling it to learn causal structures from observed mixed-type data [20]. However, hybrid methods generally require to integrate the logic and processes of multiple approaches, leading to a relatively complex algorithm design and implementation [21]. Additionally, the performance of

Wenrui Li is with the Department of Automation, Tsinghua University, Beijing, 100084, China (e-mail: li-wr23@mails.tsinghua.edu.cn).
Wei Zhang is with the School of Control Science and Engineering, Shandong University, Jinan, Shandong, 250061, China.
Qinghao Zhang is with the Department of Electrical Engineering, Tsinghua University, Beijing, 100084, China.
Xuegong Zhang is with the Department of Automation, Tsinghua University, Beijing, 100084, China.
Xiaowo Wang (*Corresponding Author*) is with the Department of Automation, Tsinghua University, Beijing, 100084, China (e-mail: xwwang@mail.tsinghua.edu.cn).

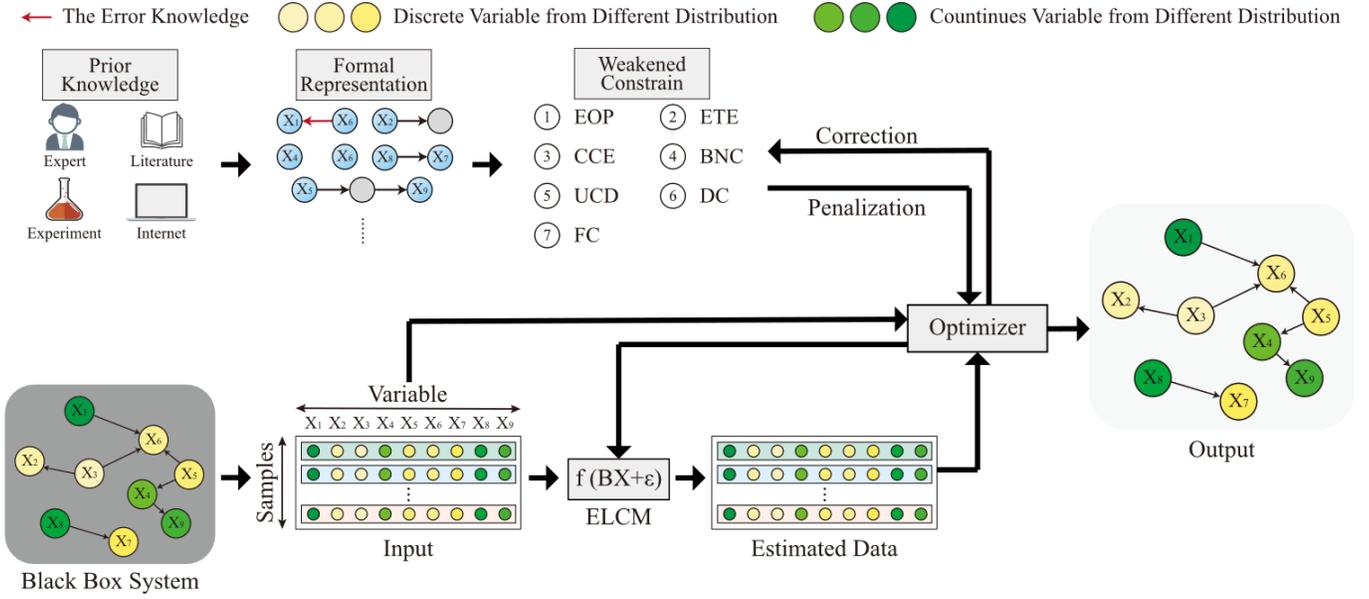

**Fig. 1**. The framework of KEEL.

those integrated methods can disturb each other, leading to an even worse result.

Function-based methods discover DAGs using function models [10]. These methods can handle more complex causal relationships, such as non-linear and multivariable interactions with high efficiency. For instance, Noteras transforms the problem of DAG structure learning into a continuous optimization problem and thus reduces computational complexity [22]. This kind of methods has strong flexibility, expansibility, and can distinguish Markov equivalence classes, thus becoming a research hotspot [23],[24]. However, these methods' performance depends on the fitness between the selected model and data, which limits the generalization [10]. Moreover, their performance suffers under complex data characterized by high-dimensional and small-sample size which increases the risk of underfitting [25],[26],[27].

The above-mentioned causal discovery methods encounter the following common challenges. *First*, they may overlook or violate fundamental principles and known facts within the domain, affecting the credibility of conclusions. [28]. *Second*, they encounter difficulties in addressing high-dimensional and small-sample data. As the number of variables and interactions increase, the risk of false discoveries increases, presenting computational and statistical hurdles [26], [27], [29]. Meanwhile, the reliability of existing methods is significantly compromised in the limited samples, leading to inaccurate causal results and diminished robustness [30], [31].

To enhance the performance of conventional causal discover methods, knowledge-assisted methods are proposed and gain increasing attentions [32], [33], [34]. These methods utilize prior knowledge to assist causal discovery. Such knowledge may come from expert judgments, literature or experimental results. By integrating prior knowledge, knowledge-assisted methods can eliminate unreasonable assumptions, effectively reducing the search space and computational complexity in the high-dimensional scenario. For instance, if there are explicit and inalterable understandings of the causal relationships of partial variables (required direct or forbidden causal relationships), it is efficient to discard inconsistent inference from the solution set to enhance performance. In cases with limited available data, knowledge assistance can provide additional support, filling in data gaps and aiding causal discovery. However, these methods heavily rely on hard-to-obtain reliable expertise and are in the preliminary stage of converting prior knowledge into strict constraints. There is a lack a systematical definition and representation of general knowledge with varying quality. Thus, it is challenging for these methods to incorporate various knowledges which are usually fuzzy, failing to fully leveraging the potential of a wide range of knowledge to enhance causal discovery performance. Moreover, these methods are unable to adjust imprecise, uncertain or unsuitable knowledge, thereby constraining flexibility and algorithmic performance. Despite demonstrating the potential of causal discovery through the combination of knowledge and data, these methods struggle to circumvent the tradeoff between high expertise dependency and the intricacies of causal discovery in complex systems.

*B. Contribution*

To bridge the gaps identified in current causal discovery methods, we propose a weakly-supervised fuzzy knowledge-data co-driven causal discovery method named KEEL (Fig. 1). KEEL synergistically integrates fuzzy knowledge and data, effectively improving the accuracy and computational efficiency of causal discovery in complex systems while reducing the dependency on expertise. KEEL adopts a novel fuzzy knowledge schema that implements weakened knowledge constraints from fuzzy knowledge. This schema supports knowledge correction and supplementation thorough interaction with data, thus allowing diverse knowledge types with varying quantities, purity, precision and certainty levels (even limited and error-prone) to guide the causal discovery process, significantly expanding the generalization. In addition, the utilization of ELCM in KEEL enables adaptability to diverse data types in the real world. In-depth theoretical analysis demonstrates the efficacy of KEEL, highlighting its superior theoretical reliability compared to traditional methods. Experiments with synthetic data demonstrate that, even with increased system complexity and a reduction in the quality of knowledge, KEEL maintains high levels of accuracy and robustness. This is particularly evident for high-dimensional, small-sample dataset. Additional validation on the real protein signal transduction dataset further highlights the advantages of KEEL in practical applications. The main contributions of this article are as follows.

1) We propose a novel formal definition and representation of fuzzy causal knowledge, maximizing the utility of available knowledge with varying quality. This approach formalizes 7 common types of fuzzy knowledge. To our best knowledge, it is the first time to systematically define the fuzzy causal knowledge formalization. It provides the theoretical basis for weakly supervised causal discovery which incorporate fuzzy knowledge to enhance accuracy, robustness, and computational efficiency in complex systems.
2) We propose weakened constraints to incorporate different types of fuzzy knowledge. This enables the weakly supervised causal discovery which is characterized by the simultaneous optimization of discovering DAGs from observational data and refining fuzzy knowledge, thus reducing the reliance of expertise.
3) We propose an extended linear causal model (ELCM) with identifiability proof as well as its corresponding optimization as the center of KEEL to capture causality patterns of data. This model is adaptable to mixed discrete and continuous data with various distributions in the presence of latent variables, thus extending the identifiable space and application scales of KEEL.

## II. METHODOLOGY

In this section, we offer the theoretical motivation and formal definition of KEEL. We begin by introducing the ELCM as the cornerstone of KEEL. The incorporation of ELCM enables KEEL to conduct causal discovery in scenarios involving multi-distribution and incomplete data. Furthermore, we formalize 7 distinct common types of fuzzy knowledge and convert them into weakened constraints.

### A. Extended Linear Causal Model

For simplification, we consider the case where all discrete variables are binary, and make the following assumptions:

**Assumption 1**: The generation mechanism of all research variables is faithful to a weighted DAG $G$, where nodes represent the research variables and edges represent their causal relationships.

**Assumption 2**: Continuous variable $X_i$ has a linear relationship with its parents $X_{PA_i}$ and latent variable $\varepsilon_i$ following any distribution $P(\varepsilon_i)$. The latent variables are continuous and mutually independent., i.e.:

$$X_i := \beta_i^T X_{PA_i} + \varepsilon_i, \quad \varepsilon_i \sim P(\varepsilon_i). \quad (1)$$

**Assumption 3**: Discrete variable $X_i$ takes the value of 1 when the linear function of its parent $X_{PA_i}$ and any distributed latent variable $\varepsilon_i$ exceeds 0. The latent variables are continuous and mutually independent., i.e.:

$$X_i := \begin{cases} 1, \beta_i^T X_{PA_i} + \varepsilon_i > 0 \\ 0, \beta_i^T X_{PA_i} + \varepsilon_i \leq 0 \end{cases}, \quad \varepsilon_i \sim P(\varepsilon_i). \quad (2)$$

**Definition 1:** The data generation model fulfilling these assumptions is called the extended linear causal model (ELCM), i.e.:

$$X_i := f_i(\beta_i^T X_{PA_i} + \varepsilon_i), \quad \varepsilon_i \sim P(\varepsilon_i), \quad (3)$$

where symbol $:=$ denotes a one-way relationship. $X_{PA_i} = [X_{PA_{i1}}, X_{PA_{i2}}, \ldots, X_{PA_{ip}}]^T$ is the p-dimension parent variables of $X_i$, and $\beta_i$ is a $p \times 1$ vector quantifying the direct causal effect of $X_{PA_i}$ on $X_i$. $\varepsilon_i$ represents the continuous latent variable which contributes to $X_i$, following any distribution $P(\varepsilon_i)$. If $X_i$ is continuous, $f_i$ is the identity function. If $X_i$ is discrete, $f_i$ is a step-like function with a zero value at the threshold.

Based on this model, we propose an objective function to minimize:

$$f(\boldsymbol{B}, \boldsymbol{\theta}) = -\ln L(\boldsymbol{B}, \boldsymbol{\theta}|X) = -\ln p(X|\boldsymbol{B}, \boldsymbol{\theta}),$$

where

$$p(X|\boldsymbol{B}, \boldsymbol{\theta}) = \prod_{i=1}^{D} p_{bi}(X_i|X_{PA_i}, \boldsymbol{\theta}_{\varepsilon_i})^{z_i} p_{ci}(X_i|X_{PA_i}, \boldsymbol{\theta}_{\varepsilon_i})^{1-z_i}$$

$$= \prod_{i=1}^{D} \prod_{n=1}^{N} \left(1 - F_{bi}(-\beta_i^T x_{PA_{in}}|\boldsymbol{\theta}_{\varepsilon_i})\right)^{x_{in} z_i} \cdot$$

$$F_{bi}(-\beta_i^T x_{PA_{in}}|\boldsymbol{\theta}_{\varepsilon_i})^{(1-x_{in})z_i} \cdot$$

$$\left(f_{ci}(x_{in} - \beta_i^T x_{PA_{in}}|\boldsymbol{\theta}_{\varepsilon_i})\right)^{1-z_i}, \quad (4)$$

$\boldsymbol{B}$ is the adjacency matrix of the weighted DAG. $\boldsymbol{\theta}$ is the parameter for the latent variable probability function. $x_{in}$ is the $n$-$th$ sample of variable $X_i$. N is the sample size. $p_{bi}$ and $p_{ci}$ represent the probability and probability density functions for the discrete and continuous variables, respectively. $z_i$ is an indicator variable, assigning 1 if $X_i$ is a discrete variable and 0 otherwise. $F_{bi}$ and $f_{ci}$ represent the probability distribution function and the density function of the latent variables for discrete and continuous variables, respectively.

Based on (4), we solve the following optimization problem to ensure sparsity and acyclicity:

$$\min_{\boldsymbol{B}} \quad L_\Lambda(\boldsymbol{B}, \boldsymbol{\theta}) = f(\boldsymbol{B}, \boldsymbol{\theta}) + \|\Lambda \cdot \boldsymbol{B}\|_1, \quad (5)$$

subject to

$$h(\boldsymbol{B}) = 0.$$

Here, $h(\boldsymbol{B})$ ensures that the directed graph corresponding to $\boldsymbol{B}$ is acyclic. $\Lambda$ is the regularization parameter matrix which allows distinct parameters for different edges. $l_1$-regularization $\|\Lambda \cdot \boldsymbol{B}\|_1 = \|\text{vec}(\Lambda \cdot \boldsymbol{B})\|_1$ is applied to avoid overfitting. Notably, the element $b_{i,j}$ of $\boldsymbol{B}$ quantifies the direct causal effect of $X_i$ on $X_j$ assuming there are no confounding variables. $b_{i,j} = 0$ indicates the absence of a causal relationship, implying no edge from $X_i$ to $X_j$. Conversely, $b_{i,j} \neq 0$ suggests the potential existence of a causal relationship, indicating an edge from $X_i$ to $X_j$. Following the optimization method proposed by Zheng et al. [22], we employ augmented Lagrange method to convert (5) into the unconstrained function (6). The L-BFGS algorithm is then used to solve it [35]. Here, $\rho$ is penalty parameter and $\alpha$ is Lagrange multiplier. Due to the limitations of machine precision, $b_{i,j}$ is not precisely zero [22]. Therefore, we introduce a threshold $\tau$ and set $b_{i,j}$ to zero for values smaller than $\tau$ in absolute terms.

$$\min_B \max_\alpha L_{\Lambda,\rho}(B, \theta, \alpha) = f(B, \theta) + \|\Lambda \cdot B\|_1 + \alpha\, h(B) + \frac{\rho}{2}|h(B)|^2. \tag{6}$$

When the prior knowledge is available, we replace $\|\Lambda \cdot B\|_1$ by a prior causal knowledge constraint $L_{\text{know}}$ (see Section II.B). Thus, Equation (6) can be reformulated as:

$$\min_B \max_\alpha L_{\Lambda,\rho}(B, \theta, \alpha) = f(B, \theta) + L_{\text{know}} + \alpha\, h(B) + \frac{\rho}{2}|h(B)|^2. \tag{7}$$

For the proof of causal direction identifiability of ELCM as well as the Optimization details of KEEL, please see Supplementary Materials.

*B. Incorporate Fuzzy Causal Knowledge*

Knowledge about causal relationships is generally fuzzy, with fuzzy knowledge easier to acquire but more difficult to process. To circumvent this tradeoff, we propose the fuzzy knowledge schema. Specifically, we formulate various types of fuzzy causal knowledge and the corresponding weakened constraints, thus incorporating knowledge which may be uncertain or imprecise to guide the causal discovery process. These constraints can be applied individually or in combination based on task requirements and available knowledge. Thus, we can specify $L_{\text{know}}$ in (7) to form the constrained optimization.

**Definition 2** [36]: For a given universe of discourse X, a fuzzy set $A$ is defined as:
$$A = \{(x, \mu_A(x)) | x \in X\}$$
where $\mu_A: X \to [0,1]$ is the membership function of $A$. $\mu_A(x)$ denotes the degree of membership of $x$ in $A$.

**Definition 3**: $\Psi = \{\Psi_i\}_{i \in I}$ and $\Omega = \{\Omega_j\}_{j \in J}$ are two non-empty sets, and $\tilde{P}(\Omega)$ is the fuzzy power set of $\Omega$. Fuzzy causal function $F$ from $\Psi$ to $\Omega$ is the mapping of $\Psi$ into $\tilde{P}(\Omega)$ such that for $\forall (\Psi, \Omega) \in \Psi \times \Omega$, $\mu_{F(\Psi)}(\Omega) = \mu_F(\Psi, \Omega)$, where $\mu_F(\Psi, \Omega)$ is the membership of $\Psi$ causing $\Omega$.

**Corollary 1**: $\mu_F(\Psi, \Omega)$ is the degree to which the relation of $\Psi$ to $\Omega$ satisfies the causality property in research domains. $\mu_F(\Psi, \Omega) = 0$ implies the relation from $\Psi$ to $\Omega$ is exactly non-causal, while $\mu_F(\Psi, \Omega) = 1$ implies the relation from $\Psi$ to $\Omega$ is exactly causal. Considering $0 < \mu_F(\Psi, \Omega) < 1$, $\mu_F(\Psi, \Omega) > 1 - \mu_F(\Psi, \Omega)$, i.e. $\mu_F(\Psi, \Omega) > 0.5$ implies the relation from $\Psi$ to $\Omega$ is not exactly causal, while $\mu_F(\Psi, \Omega) < 0.5$ implies the relation from $\Psi$ to $\Omega$ is not exactly non-causal. Otherwise, $\mu_F(\Psi, \Omega) = 0.5$ implies no constructive opinions about causal relation from $\Psi$ to $\Omega$, indicating maximum fuzziness.

**Definition 4**: Assume each fuzzy causal knowledge correspond to a fuzzy causal mechanism $Q$. $Q$ is a quintuple $(\mathbf{P}, \mathbf{C}, F, \mathbf{M}, \mathbf{L})$, where $\mathbf{P}$ is a non-empty set of parent variables, $\mathbf{C}$ is a non-empty set of child variables, $F$ is the fuzzy causal function, $\mathbf{M}$ is the set of the mediating variables in $F$, and $\mathbf{L}$ is the corresponding set of the mediation steps.

**Notation 1**: In this paper, we denote a fuzzy causal knowledge as K and the corresponding fuzzy causal mechanism as $Q_K$. The variable representing entity T is denoted as $V_T$. Supposing $S$ is the set of statements implying fuzzy causal knowledge, $F$ from $\Psi$ to $\Omega$ is denoted as $F_{\Psi \xrightarrow{S} \Omega}$. The crisp set of research variables $N$, and the target variables $X$ and $Y$ are denoted as **N**, **X** and **Y**, respectively. Both **X** and **Y** are subsets of **N**.

**Example 1**: K: The promotion on Black Friday stimulates consumption.

$Q_K: \mathbf{P} = \{V_{\text{promotion}}\}$, $\mathbf{C} = \{V_{\text{consumption}}\}$, $S = \{\text{stimulate}\}$,

$F = F_{\mathbf{P} \xrightarrow{S} \mathbf{C}}$, $\mu_F(V_{\text{promotion}}, V_{\text{consumption}}) > 0.5$, $\mathbf{M} = \emptyset$, $\mathbf{L} = \emptyset$.

For more examples of Definition 4-4.5, please see Supplementary Materials.

**Definition 4.1**: A fuzzy causal knowledge $K_1$ is called **EOP** (exposure and outcome property) if it corresponds to a $Q_{K1}$ such that:

$\mathbf{P} = \mathbf{N}$, $\mathbf{C} = \mathbf{X}$, $\mu_F(N, X) < 0.5$ or $\mathbf{P} = \mathbf{X}$, $\mathbf{C} = \mathbf{N}$, $\mu_F(X, N) < 0.5$;

$\mathbf{M} = \emptyset$, $\mathbf{L} = \emptyset$.

**Definition 4.2**: A fuzzy causal knowledge $K_2$ is called ETE (end to end) if it corresponds to a $Q_{K2}$ such that:

$$\mathbf{P} = \mathbf{X}, \mathbf{C} = \mathbf{Y}, \mu_F(X, Y) \approxeq 1,$$
$$\mathbf{M} = \{(N, \mu_\mathbf{M}(N)) | N \in \mathbf{N}\},$$
$$\mathbf{L} = \{((N_i, N_j), \mu_\mathbf{L}((N_i, N_j))) | N_i \in \mathbf{N}, N_j \in \mathbf{N}, i \neq j\}.$$

**Definition 4.3**: A fuzzy causal knowledge $K_3$ is called CCE (conditional cause and effect) if it corresponds to a $Q_{K3}$ such that:

$$\mathbf{P} = \mathbf{X}, \mathbf{C} = \mathbf{Y}, \mu_F(X, Y) > 0.5, \mathbf{M} = \emptyset, \mathbf{L} = \emptyset$$

**Remark 1**: On the one hand, a conditional cause is a potential cause that has a lower likelihood of leading to a specific outcome. On the other hand, relationships that supported by such knowledge are more likely to be causal compared to those lack knowledge assurance. Conditional causality is often observed in situation where plausible causal networks or subnetworks provide a broad understanding of causality, even though there may be exceptions or oversights in certain cases. For example, causal knowledge derived from a knowledge graph can effectively demonstrate conditional causality.

**Definition 4.4:** A fuzzy causal knowledge $K_4$ is called BNC (basically noncausal) if it corresponds to a $Q_{K4}$ such that:

$$\mathbf{P} = \mathbf{X}, \mathbf{C} = \mathbf{Y}, \mu_F(X, Y) < 0.5,$$
$$\mathbf{M} = \{(N, \mu_\mathbf{M}(N)) | N \in \mathbf{N}\},$$
$$\mathbf{L} = \{((N_i, N_j), \mu_\mathbf{L}((N_i, N_j))) | N_i \in \mathbf{N}, N_j \in \mathbf{N}, i \neq j\}.$$

**Remark 2**: This knowledge is a belief that two target variables is basically noncausal. With this belief, it is preferable to exclude potential spurious causality, such as confounding effects and data selection bias. It indicates unnecessary or incorrect inferences about causal mechanisms.

**Definition 4.5:** A fuzzy causal knowledge $K_5$ is called UCD (unknown causal direction) if it corresponds to a $Q_{K5}$ such that:
$$P=\{(X,\mu_P(X)),(Y,\mu_P(Y))|X\in \mathbf{X}, Y\in \mathbf{Y}\},$$
$$C=\{(X,\mu_C(X)),(Y,\mu_C(Y))|X\in \mathbf{X}, Y\in \mathbf{Y}\},$$
$$\mu_P(X)+\mu_C(X)=1, \mu_P(Y)+\mu_C(Y)=1,$$
$$\mu_P(X)+\mu_P(Y)=1, \mu_F(X,Y)=1,$$
$$\mathbf{M}=\emptyset, \mathbf{L}=\emptyset.$$

**Remark 3**: This knowledge gives two causal connected entity $X$ and $Y$, but provide insufficient evidence or no preference for $X$ cause $Y$ or vice versa. It indicates the cause-and-effect pairs to be further investigated, reducing the likelihood of no-correlation and spurious causality due to confounder.

**Remark 4**: To generalize the fuzzy causal knowledge and mechanism pairs $(K, Q_K)$, we can also represent the crisp knowledge about required direct or forbidden causal relationships in our fuzzy schema. A causal knowledge about direct causal relationships (**DC**) from $\mathbf{X}$ to $\mathbf{Y}$ can be interpreted as a fuzzy causal knowledge $K_6$ with $Q_{K6}=(\mathbf{X},\mathbf{Y},F,\emptyset,\emptyset)$ where $\mu_F(X,Y)=1$. A causal knowledge about forbidden causal relationships (**FC**) from $\mathbf{X}$ to $\mathbf{Y}$ can be interpreted as a fuzzy causal knowledge $K_7$ with $Q_{K7}=(\mathbf{X},\mathbf{Y},F,\emptyset,\emptyset)$ where $\mu_F(X,Y)=0$.

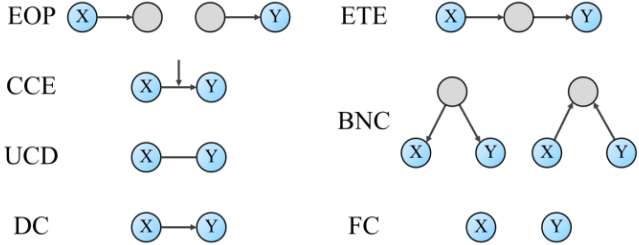

**Fig. 2**. DAG representation of fuzzy causal knowledge

We incorporate these 7 types of fuzzy causal knowledge (Fig. 2) as weakened constraints in the causal discovery process. This enables the complementary integration of knowledge and data. Thus, fuzzy knowledge helps to improve accuracy and computational efficiency, while data helps to correct and supplement knowledge.

We denote the row/column index of node $x$ as $i(x)$. We denote $|\mathbf{B}|^1=\|\text{vec}(\mathbf{B})\|_1, |\mathbf{B}|^2=\|\mathbf{B}\|_F^2$.

**EOP**: In the causal DAG, this fuzzy knowledge indicates that the node $x$ representing target variable $X$ tends to have no ancestors, or the node $y$ representing target variable $Y$ tends to have no descendants. Therefore, we penalize the in-degree of $x$ and the out-degree of $y$, making the corresponding edge weights approach zero. The penalty function can be articulated as follows:
$$|\mathbf{B}_{*,i(x)}|^\alpha+|\mathbf{B}_{i(y),*}|^\alpha, \alpha=1,2.$$

**ETE**: In the causal DAG, this fuzzy knowledge indicates that node $y$ is reachable from node $x$, but the hops, topological order of hops, and paths from $x$ to $y$ are unclear. Therefore, we constrain the condition holds: there exists at least a $k$-length path from $x$ to $y$, where $k<D$ and $D$ is the number of nodes. For the graph's binary adjacency matrix $\mathbf{A}$, $(\mathbf{A}^k)_{i(x),i(y)}$ equal to the number of $k$-length paths from $x$ to $y$. Thus, such condition holds if and only if $\sum_{k=0}^{D-1}(\mathbf{A}^k)_{i(x),i(y)}>0$. To generalize this to adjacency matrix $\mathbf{B}$ with both positive and negative values and simplifies the power sum, we derive the constraint:
$$(I+\mathbf{B}\bullet\mathbf{B})^{D-1}{}_{i(x),i(y)}>0,$$
and the penalty function:
$$|\min(0,(I+\mathbf{B}\bullet\mathbf{B})^{D-1}{}_{i(x),i(y)}-s)|^\alpha, \alpha=1,2.$$
where $s$ is the slack variable, $s>0$. We discover the paths from $x$ to $y$ and the rest of the graph structure fitting the data pattern in the constraint domain.

**CCE**: In the causal DAG, this fuzzy knowledge indicates that the edge from node $x$ to $y$ is more likely to exist but lacks confidence. On the one hand, compared to general potential edges, such an edge is more likely to be identified, so we push its weight away from zero. On the other hand, there lacks evidence supporting this high likelihood of existence, so we push its weight towards zero. For sparsity, all potential edges' existence has an $l_1$-penalty with a penalty coefficient of $\omega_1$. Since the existence of such an edge is based on a higher likelihood, its penalty should be smaller than that for general potential edges. Therefore, edges supported by **CCE** are assigned a smaller penalty coefficient than those without such support. The penalty function corresponding to this trade-off is:
$$|\mathbf{B}_{i(x),i(y)}|^1-\beta\max(0,|\mathbf{B}_{i(x),i(y)}|^1),$$
where $0<\beta\leq 1, \beta\approx 1$. Namely, the penalty function is:
$$\frac{1}{\gamma}|\mathbf{B}_{i(x),i(y)}|^1,$$
where $\gamma\gg 1$.

**BNC**: In the causal DAG, this fuzzy knowledge indicates that node $y$ is not reachable from node $x$, yet it still potentially considers that node $y$ is reachable from node $x$. Therefore, we make a constraint:
$$(I+\mathbf{B}\bullet\mathbf{B})^{D-1}{}_{i(x),i(y)}=0,$$
with tolerance to violation. The penalty function is:
$$|(I+\mathbf{B}\bullet\mathbf{B})^{D-1}{}_{i(x),i(y)}|^\alpha, \alpha=1,2.$$
It tends to exclude the possibility that there exists a path from node $x$ to $y$. Thus, it helps to distinguish weakly connected (confounding structure ($x\leftarrow n\rightarrow y$), colliding structure ($x\rightarrow n\leftarrow y$)) and unconnected structure ($x\quad n\rightarrow y$) from path ($x\rightarrow n\rightarrow y$). It can narrow down solution space and reduce unexpected discovered paths from observational data. Meanwhile, it allows for emerging paths countering to common sense but with strong data support.

**UCD**: In the causal DAG, this fuzzy knowledge implies the presence of an edge with node $x$ and $y$ as endpoints, but its direction is uncertain. Therefore, we constrain that there exists an edge between node $x$ and $y$, regardless of its direction. Namely, the constraint is that the edge weight $b_{i(x),i(y)}$ and

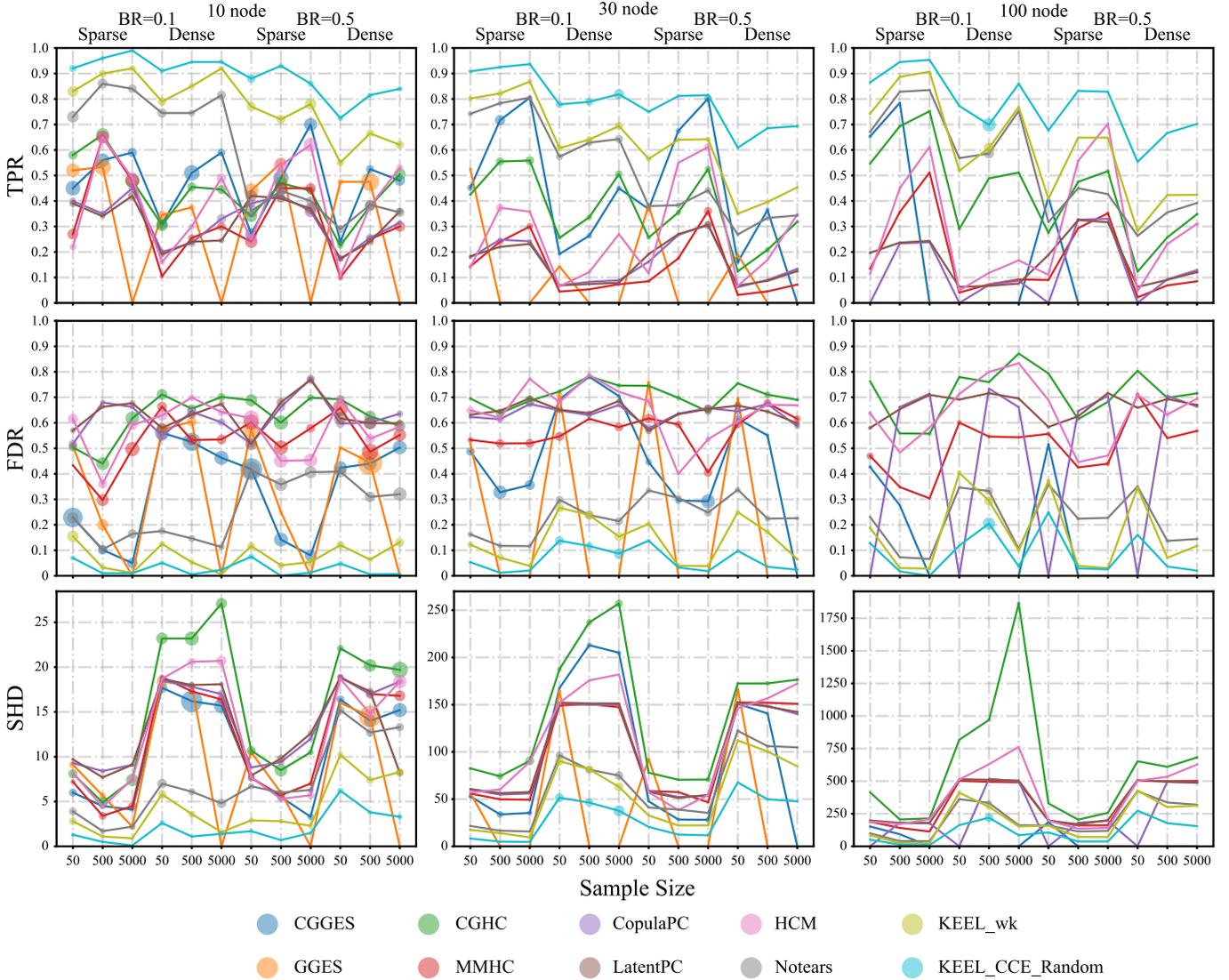

**Fig. 3**. The performance of different methods in Case A, Dataset1. The evaluation considers varying node numbers, average degree, binary ratio (BR), and sample size. The circle size represents the relative magnitude of the standard deviation.

$b_{i(y),i(x)}$ cannot both be zero:

$$c = 0,$$

where

$$c = \begin{cases} 1, if\ |b_{i(x),i(y)}| = 0\ and\ |b_{i(y),i(x)}| = 0 \\ 0, otherwise \end{cases}.$$

The penalty function is:

$$|[\max(0, \tau - |\boldsymbol{B}|) \bullet \max(0, \tau - |\boldsymbol{B}^T|)]_{i(x),i(y)}|^\alpha, \alpha = 1,2,$$

where $\tau$ is the threshold, $\tau > 0$. This penalty function being equal to zero stipulates that at least one of the following is true: the edge weight from node $x$ to $y$ surpasses the threshold, or the edge weight from node $y$ to $x$ does so. Notably, the acyclicity constraint in (5) prevents a mutual causal loop between node $x$ and node $y$. It implies that bi-direction cannot hold. Therefore, combined with the acyclicity constraint, this penalty tries to ensure there is one and only one directed edge. It pays more attention to putative edges, while using data to conduct refinement, test or correction. Thus, it enables simultaneous local causal direction identification of targeted bivariate pairs. Moreover, it can embed partially modifiable graph skeleton. Thus, it aids in reducing solution space, decreasing omissions of inferred edges, and can be used in finetuned causal discovery with putative multivariate graph skeletons.

**DC**: The direct edge from node $x$ to $y$ can be explicitly determined. It can be bounded that the edge weight from node $x$ to node $y$ is greater than the threshold:

$$\boldsymbol{B}_{i(x),i(y)} \geq \tau.$$

**FC**: It is explicitly indicated that there is no direct edge from node $x$ to $y$. It can be bounded that the edge weight from node $x$ to $y$ is zero:

$$0 \leq \boldsymbol{B}_{i(x),i(y)} \leq 0.$$

### III. RESULTS AND DISCUSSIONS

In this section, we validate the effectiveness and advantages of KEEL through a series of comparative experiments, followed by the comparison with eight state-of-the-art causal discovery methods on synthetic datasets simulating different scenarios, including high-dimensional and small-sample situations and different data distributions. We also investigate the impact of knowledge constraints on improving global accuracy and local interpretability, highlighting the advantages of utilizing fuzzy knowledge. In addition, we demonstrate that integrating diverse types of knowledge constraints can further enhance the performance, suggesting that all types of knowledge provide unique benefits without redundance. Moreover, the robustness of KEEL is assessed under varying levels of knowledge noise and quantity. Finally, we validate KEEL on real bioinformatics data, and discover the protein signaling transduction network.

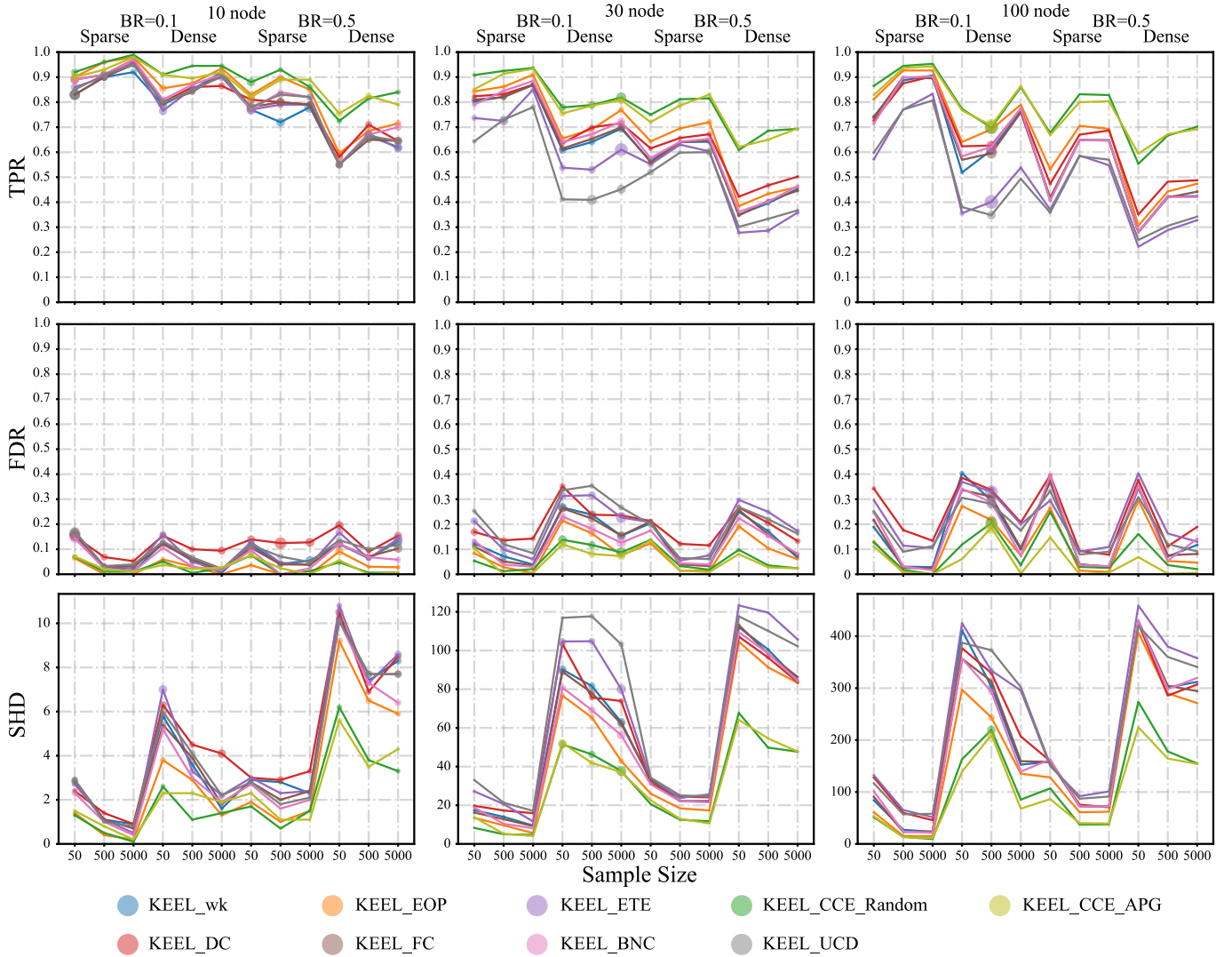

**Fig. 4**. The performance of different types of weakened constraints in Case B, Dataset 1. The evaluation considers varying node numbers, average degree, binary ratio (BR), and sample size. The circle size represents the relative magnitude of the standard deviation.

*A. Benchmarking Methods and Evaluation Metrics*

We perform a comparison between KEEL and 8 benchmark methods. These include constraint-based methods (CouplaPC and LatentPC), score-based methods (CGGES, GGES, and CGHC), hybrid methods (HCM and MMHC) and a function-based method (Notears). We evaluate KEEL against MMHC which learns discrete CPDAGs by discretizing continuous variables and selecting the optimal DAG from the Markov equivalence class. Additionally, we compare KEEL with Notears which learns continuous DAGs by modeling discrete variables as Gaussian distributions. The other 6 methods are capable of handling mixed data types.

The accuracy of causal discovery is evaluated using the following metrics: true positive rate (TPR), false discovery rate (FDR), and structure hamming distance (SHD). TPR and FDR values range between 0 and 1. Higher TPR, along with lower FDR and SHD indicate improved accuracy. For datasets generated from DAGs with 10 and 30 nodes, the average metrics are calculated over 10 random replicates. For datasets derived from DAGs with 100 nodes, these metrics are averaged over 3 random repetitions. Both the mean values and standard deviations for those metrics are calculated.

*B. Synthetic Datasets with Various Knowledge Type*

We adopt the ER model for random DAG generation, sampling edges independently and uniformly without replacement. The edge weights follow uniform distributions within the intervals [-2, -0.5] and [0.5, 2]. This procedure yields weighted DAGs and their corresponding adjacency matrix $B$.

Table 1. The improvement of global accuracy and local interpretability for adding ETE and UCD

| Scenario | | | ETE | | | UCD | | |
|---|---|---|---|---|---|---|---|---|
| | | | FDR | TPR | CP Prop | FDR | TPR | CD Prop |
| BR 0.1 | Sparse | N50 | 0.287 ± 0.165 | 0.841 ± 0.142 | 0.524 ± 0.382 | 0.287 ± 0.165 | 0.912 ± 0.078 | 0.771 ± 0.343 |
| | | N500 | 0.025 ± 0.043 | 0.950 ± 0.050 | 1.000 ± 0.000 | 0.025 ± 0.043 | 0.967 ± 0.047 | 1.000 ± 0.000 |
| | | N5000 | 0.150 ± 0.050 | 0.850 ± 0.050 | 1.000 ± 0.000 | 0.150 ± 0.050 | 0.950 ± 0.050 | 1.000 ± 0.000 |
| | Dense | N50 | 0.122 ± 0.070 | 0.653 ± 0.112 | 0.704 ± 0.328 | 0.122 ± 0.070 | 0.900 ± 0.050 | 0.671 ± 0.303 |
| | | N500 | 0.055 ± 0.066 | 0.706 ± 0.103 | 0.830 ± 0.154 | 0.055 ± 0.066 | 0.914 ± 0.113 | 0.893 ± 0.182 |
| | | N5000 | 0.058 ± 0.055 | 0.713 ± 0.100 | 0.750 ± 0.250 | 0.058 ± 0.055 | 0.936 ± 0.051 | 0.809 ± 0.350 |
| BR 0.5 | Sparse | N50 | 0.149 ± 0.099 | 0.763 ± 0.116 | 0.709 ± 0.378 | 0.149 ± 0.099 | 0.800 ± 0.176 | 0.648 ± 0.355 |
| | | N500 | 0.032 ± 0.053 | 0.819 ± 0.079 | 0.696 ± 0.301 | 0.032 ± 0.053 | 0.889 ± 0.110 | 0.815 ± 0.266 |
| | | N5000 | 0.059 ± 0.059 | 0.819 ± 0.088 | 0.577 ± 0.434 | 0.059 ± 0.059 | 0.920 ± 0.108 | 0.850 ± 0.320 |
| | Dense | N50 | 0.163 ± 0.097 | 0.522 ± 0.094 | 0.462 ± 0.168 | 0.163 ± 0.097 | 0.838 ± 0.042 | 0.762 ± 0.119 |
| | | N500 | 0.113 ± 0.077 | 0.631 ± 0.140 | 0.393 ± 0.190 | 0.113 ± 0.077 | 0.833 ± 0.085 | 0.686 ± 0.146 |
| | | N5000 | 0.160 ± 0.119 | 0.610 ± 0.126 | 0.554 ± 0.276 | 0.160 ± 0.119 | 0.925 ± 0.075 | 0.750 ± 0.250 |

Cp Prop refers to the proportion of iterations that correctly identify complete paths when traversing the path endpoints that are overlooked without applying the ETE constraint. CD Prop refers to the proportion of correctly identified causal directions in skeletons that are overlooked without applying the UCD constraint. This table reports the 'mean +/- standard deviation' of performance metrics across 10 runs on 10 nodes DAGs.

Each column vector in $B$ corresponds to a variable $X_i$, of which non-zero weight coefficients indicate the parent variables of $X_i$. For the generated DAGs, we create Dataset 1 and Dataset 2 via ELCM. For scenarios that require a knowledge graph as the fuzzy knowledge source in Dataset 1, we simulate synthetic knowledge graphs.

1) **Dataset 1**

To comprehensively evaluate the effectiveness of various causal discovery methods, we simulate DAGs with varying node numbers and average degree, as well as the corresponding datasets with varying binary ratio and sample size.

To investigate the impact of increasing node numbers on performance, we simulate DAGs with 10, 30, and 100 nodes, respectively. To evaluate how different levels of edge sparsity affect performance, we consider two scenarios of sparse and dense DAG. For sparse DAGs, the average degree is set to 2 for D=10, and 4 for both D=30 and D=100. For dense DAGs, it is set to 4 for D=10 and 10 for D=30 and D=100. The latent variables in this study are generated from the standard Gaussian distribution. We vary the proportion of discrete nodes to 0.1 and 0.5. The sample sizes are set as 50, 500, and 5000 to evaluate performance under conditions of data scarcity and sufficiency. We construct partial knowledge graphs with noise. Positive edges for simulating knowledge are randomly selected from the true edge set, with probabilities ranging from 0.1 to 0.9. False positive edges for simulating noise are randomly selected with probabilities between 0.01 and 0.1. This synthetic knowledge graph is referred to as a **random noisy knowledge graph** and serves as the default setting. Additionally, considering that relationships with stronger causal effects are more likely to be established, we further design sampling strategy based on edge weights. In details, the weight interval [0.5, 2] is transformed to the interval [-2.5, 2.5], and a sigmoid function is used to derive the sampling probabilities for the positive edges. This synthetic knowledge graph is termed as **APG (absolute probability sampling generated) noisy knowledge graph**, designed for comparative analysis with the random noisy knowledge graph.

On Dataset 1, we consider the following cases and conduct experiments.

**Case A**: KEEL-without knowledge (denoted as KEEL_wk) is compared with benchmark methods that lack prior knowledge. Additionally, by incorporating **CCE** constraints from a random noisy knowledge graph, we demonstrate the benefits of integrating fuzzy knowledge.

**Case B**: We compare KEEL_wk with all seven constraints mentioned above to analyze the impact of different knowledge constraints on global accuracy, including the cases of a random noisy knowledge graph and an APG noisy knowledge graph. Additionally, we investigate the impact of knowledge constraints on improving local interpretability. This is achieved by highlighting the efficacy of the **ETE** constraint in identifying causal chains and the **UCD** constraints in in identifying cause and effect.

**Case C**: We generate **EOP** and **CCE** and evaluate the performance of KEEL under individual and combined constraints.

**Case D**: We generate random noisy knowledge graphs of various sizes by sampling true edges at 10%, 30%, 50%, 70%, and 90%. Simultaneously, we simulate noise by sampling different ratios of edges from the set of non-true edges (1%, 2%, 5%, and 10%). This is used to compare the effect of different noise levels on the performance of KEEL under the **CCE** constraint.

2) **Dataset 2**

Similar to Dataset 1, we generate DAGs with 30 nodes with varying sparsity and the corresponding datasets with varying binary ratios and sample sizes. Based on this, we adjust the distributions of the latent variables, assigning the matched or mismatched distribution assumptions to KEEL_wk to assess the its scalability.

**Gaussian-Gaussian**: All latent variables follow a Gaussian distribution. *First*, we test the performance of KEEL with matched assumptions. *Second*, we test the performance with mismatched assumptions that continuous variables are generated from Laplace latent variables, while discrete variables come from Logistic latent variables.

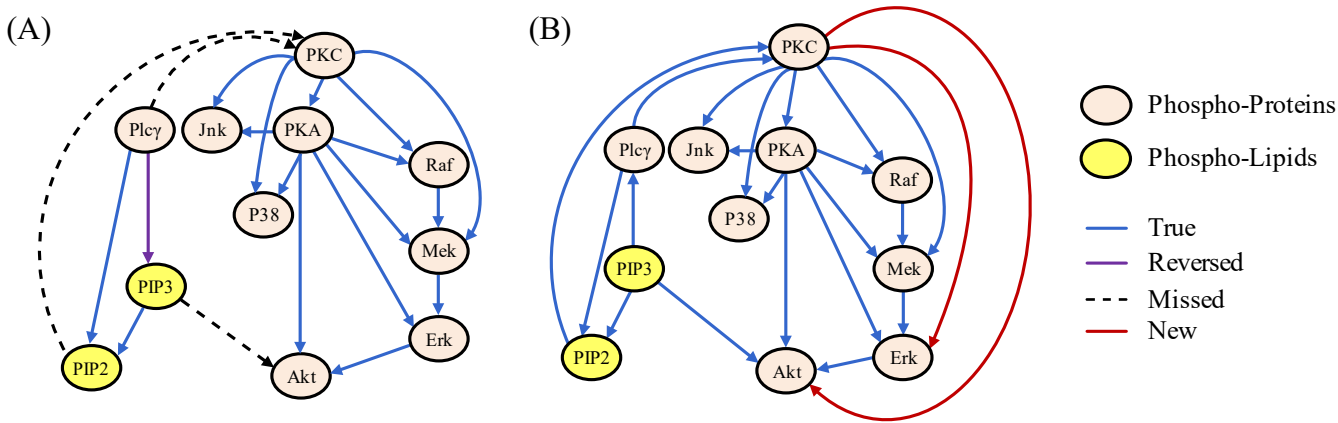

**Fig. 5**. Discovered protein signal transduction network. (A) The original result derived from the original dataset. The PIP3-Plcr connection is mistakenly identified in the opposite direction. (B) The results obtained by the KEEL_CCE on the observational data.

**Cauchy-Gaussian**: Cauchy distributed latent variables for the continuous variables and Gaussian distributed latent variables for the discrete variables. *First*, we test the performance of KEEL with the matched assumptions. *Second*, we test the performance with the mismatched assumptions that all the variables generated from Gaussian distributed latent variables.

*C. Results for Synthetic Datasets*

1) **Results of Case A of Dataset 1**

From Fig. 3, it is evident that KEEL outperforms other methods. In high-dimensional graphs, the parent set of a node may exceed 10, and the size of the adjacency matrix may reach 100×100, with each directed edge potentially interacting with others, making the probability space vast and the DAG identification challenging for all methods. The constraint-based methods may struggle due to statistical test limitations and the score-based methods fail to distinguish Markov equivalence class and may suffer from high computational complexity. For instance, GGES fails to scale to 100 nodes and dense scenarios due to its unacceptable computational complexity. The performance of function-based methods suffers when confronted with multi-distributed incomplete data due to the model assumptions may not hold. Additionally, identifying causal graph from observed data is hard not only due to high dimensionality and limited sample size, but also because the causal identifiability is only ensured under specific conditions in theory. In contrast, KEEL can handle various types of mixed data and exhibits stable performance as the number of nodes and edges increases and the sample size decreases. Additionally, compared to KEEL_wk, KEEL with weakened knowledge constraints consistently achieve optimal performance. It indicates that the proposed knowledge constraints can effectively complement data, extend the causal identifiable space, and improve the accuracy and efficiency of causal discovery in high-dimensional and small-sample data scenarios.

2) **Results of Case B of Dataset 1**

The results in Fig. 4 highlight the effects of the different types of weakened constraints on improving the accuracy of causal discovery. Overall, the proposed knowledge constraints robustly improve the accuracy of causal discovery across various scenarios. These constraints led to an increase in TPR ranging from 2.5% to 31%, and a decrease in FDR ranging from 0.1% to 34%. Among them, the **CCE** constraint show the most significant improvement, with its advantage becoming more evident with increasing data dimensionality, decreasing sample size, and increasing binary ratios. Additionally, we find that strict constraints on direct causal relationships might interfere with the discovery of other relationships. For example, while the **FC** constraint also reduces FDR and SHD, its improvement in TPR is inconsistent. Introducing the **DC** constraint can increase TPR, especially with small to medium sample sizes, but it also carries a risk of increasing FDR and SHD. Therefore, introducing crisp knowledge does not always guarantee better performance. It may lead to overly strict and redundant constraints, undermining overall accuracy. Loosening constraints appropriately helps to fully explore the valuable information within the data, which is in line with the initial intention of introducing weakened constraints in this study.

It is worth noting that some constraints are effective in explaining local structures. For example, in low-dimensional scenarios, the **ETE** and **UCD** constraints improve TPR while FDR and SHD remained relatively stable. In medium and high-dimensional scenarios, these constraints can reduce overall accuracy, primarily due to oversampling of redundant relations and prioritizing local structure accuracy. Further study on the local interpretability of **ETE** and **UCD** find that **ETE** increases the likelihood of correctly identifying the complete path, while **UCD** helps accurately identify causal directions within multivariable undirected graph skeletons (Table 1).

Overall, we can confirm the effectiveness of all proposed constraints. It is concluded that the proposed fuzzy knowledge formalization schema enables knowledge of various precision and certainty levels to facilitate causal discovery. It reduces the reliance on expertise and expands the scope of prior information that can be utilized to enhance the accuracy. Additionally, it aids in identifying complete causal chains and cause-effect relationships within fixed graph skeletons, suggesting its applicability to a broader range of causal-related tasks.

**Table 2. Sachs Result**

| Scenario | FDR | TPR | SHD | NNZ | Scenario | FDR | TPR | SHD | NNZ |
|---|---|---|---|---|---|---|---|---|---|
| CopulaPC | 0.42 | 0.35 | 18 | 12 | Keel-EOP | 0.28 | 0.65 | 10 | 18 |
| LatentPC | 0.42 | 0.35 | 18 | 12 | Keel-ETE | 0.07 | 0.65 | 8 | 14 |
| CGGES | 0.50 | 0.35 | 20 | 14 | **Keel-CCE** | **0.09** | **1.00** | **2** | **22** |
| GGES | 0.59 | 0.45 | 22 | 22 | Keel-BNC | 0.18 | 0.70 | 9 | 17 |
| CGHC | 0.44 | 0.50 | 21 | 18 | Keel-UCD | 0.21 | 0.75 | 8 | 19 |
| MMHC | 0.47 | 0.40 | 20 | 15 | Keel-DC | 0.26 | 0.70 | 10 | 19 |
| HCM | 0.70 | 0.15 | 18 | 10 | Keel-FC | 0.24 | 0.65 | 10 | 17 |
| Notears | 0.57 | 0.30 | 15 | 14 | Sachs_observational | 0.20 | 0.40 | 12 | 10 |
| Notears-MLP | 0.63 | 0.30 | 16 | 16 | Sachs_interventional | 0.06 | 0.80 | 4 | 17 |
| Keel-wk | 0.20 | 0.60 | 10 | 15 | Keel-EOP | 0.28 | 0.65 | 10 | 18 |

3) **Results of Case C of Dataset 1**

As shown in Supplementary Fig. S1, the performance improvement gained from combining two knowledge constraints exceeds the effect of any single constraint. This suggests that different knowledge constraints contain unique information, and their combination can further enhance the performance of causal discovery.

4) **Results of Case D of Dataset 1**

The results in Supplementary Fig. S2 show that the introduction of merely 10% random uncertain knowledge in KEEL can boost TPR by up to 14% and reduce FDR by up to 17%. As the quantity of knowledge increases, the performance improvement becomes more pronounced. Specifically, the introduction of 30% knowledge strikes the best balance between performance improvement and the cost of acquiring knowledge. However, introducing a higher proportion of knowledge, such as 90%, may exhibit diminishing marginal benefits, despite achieving the best performance. Furthermore, the introduction of noisy knowledge enables robust identification of causal structures even when the signal-to-noise ratio is approximately 1, with a moderate improvement in TPR and reductions in FDR and SHD, despite potential slight effects on performance gains due to increased noise levels.

5) **Results of Dataset 2**

One advantage of KEEL lies in its theoretical ability to adapt to variables of any distribution. In the Gaussian-Gaussian case (Supplementary Table S1), the proposed method with matched assumptions outperforms that with mismatched assumption of the Laplace-Logistic distribution, which is the designated distribution in related mixed causal discovery methods [37]. In the Cauchy-Gaussian case (Supplementary Table S2), the proposed method with matched assumptions significantly outperforms that with the mismatched assumption of the Gaussian-Gaussian distribution. Related causal discovery methods like Notears assume a Gaussian distribution when dealing with continuous variables [22], [38]. These methods demonstrate that even on non-Gaussian distributed data, the mismatched assumption can still yield reasonable performance. However, we find that on data far from the Gaussian distribution, such as the heavy-tailed distributions like the Cauchy distribution, the mismatched Gaussian assumption results in decreased performance. Therefore, the ability of KEEL to effectively handle variables with various distributions provides a significant advantage in the field of causal discovery.

*D. Real-World Application*

Next, we apply KEEL to analyze the Sachs dataset [39]. This dataset consists of a validated protein signaling network with 20 connections (edges). The original interventional-observational combined dataset contains continuous measurements from 7466 single-cell samples, covering 11 protein and phospholipid variables. We extract the observational data from the original dataset and discretize the relatively balanced distributed features of PKA and PKC into binary vectors using the mean values of their distributions. This allows for inputting suitable mixed data into mixed methods for further analysis. It is worth noting that our method uses only 853 observational samples for causal discovery, whereas the study of Sachs *et al.* uses all 7,466 observational and interventional samples.

We extract a series of established connections to serve as the source for fuzzy causal knowledge. Briefly, for **EOP**, we input the knowledge that PIP3 may have no excitations, and AKT, P38, and Jnk may have no response among the 11 variables. For **ETE**, we input knowledge that Raf regulates Erk via unknown paths. For **CCE**, we input the endogenous knowledge of the noisy inference of KEEL-wk and exogenous knowledge of potential influence connections, including Raf-Mek, Mek-Erk, Plcr-PKC, PIP2-PKC, PIP3-AKT, and PKA-Raf [40], [41]. For **BNC**, we input knowledge that there may be no path from Mek to Raf. For **UCD**, we input knowledge that connections exist between Raf and PKA, PKC and Plcr, PKA and PKC, P38 and PKA. Similarly, for **DC**, we specify that there must be connections between Raf-Mek and Mek-Erk. Additionally, for **FC**, we specify that there is no connection from Mek to Raf.

In this study, we compare different methods using TPR, FDR, SHD as well as the corresponding number of estimated edges (NNZ). Considering the nonlinear relationships in real-world, we also compare KEEL with Notears-MLP which is designed for nonlinear data [42]. Notears-MLP estimates 16 edges with an SHD of 16. In contrast, KEEL without knowledge estimates 15 edges with an SHD of 10, highlighting the superiority of KEEL even without knowledge in practical applications. We further analyze the results of KEEL with fuzzy knowledge.

KEEL_CCE achieves the best performance among all the methods. (Fig. 5 and Table 2). In prior research, Sachs et al. reports predicting an undirected graph of 10 edges using the observational data, which is less consistent with the ground truth. They augment the dataset with interventional samples, resulting in a 7466-sample dataset, and improve the performance to an SHD of 4 with 17 estimated edges. In comparison, using less observational data, KEEL_CCE estimates 22 edges with an SHD of 2, accurately estimating all known connections. Moreover, all other knowledge constraints also performed relatively well on the observational data. For example, **ETE** helps identify the biologically significant Raf-Mek-Erk pathway, even outperforming the **DC** case that directly requires Raf-Mek and Mek-Erk connections. This demonstrates the advantages of weakened constraints. Additionally, **UCD** helps to identify excitations and responses in protein signal transduction, and improves the overall causal discovery performance with respect to TPR and SHD.

We further conduct an in-depth analysis of the results from KEEL_CCE. As shown in Fig. 5, despite the erroneous inclusion of the Plcr-PIP3 direction in the prior knowledge graph, KEEL_CCE successfully identifies the PIP3-Plcr connection, which is incorrectly identified in the opposite direction in the original estimation [39]. This finding confirms the robustness of the proposed method against knowledge noise. Additionally, the two new connections discovered by KEEL_CCE (PKC-Erk and PKC-AKT) have been experimentally validated in recent years, which were not identified by computational methods in previous studies [43], [44].

These results not only verify the effectiveness of the knowledge formalization schema but also demonstrate that the reasonable incorporation of even a small amount of fuzzy knowledge as weakened constraints can effectively improve causal discovery performance and reduce the dependence on interventional experiments, expertise, and data. Thus, KEEL is potential to provide new insights for future computational approaches to uncover complex biological mechanisms.

## IV. CONCLUSION

In this study, we develop a weakly-supervised fuzzy knowledge and data co-driven causal discovery method named KEEL. KEEL adopts a new fuzzy knowledge inspiration schema. This allows different amounts, purities, precision and certainty levels of knowledge to be represented as weakened constraints in the causal discovery process, effectively reducing the search space and guiding the causal discovery. This improves the accuracy and computational efficiency of causal discovery and enhances the robustness to data scarcity and knowledge error through a complementary mechanism. Additionally, the use of ELCM in KEEL extends its application to scenarios with multi-distribution and incomplete data. We also integrate causal discovery based on observational data, knowledge revision, and supplementation into a continuous optimization task to achieve more reliable identification of causal relationships. Experiments demonstrate that KEEL can accurately and robustly identify causal relationships in complex systems, characterized by high-dimensional and small-sample size, while reducing the reliance on expertise.